\documentclass[a4paper,11pt]{article}

\usepackage{breakurl}
\usepackage[breaklinks=true]{hyperref}
\usepackage{breakcites}

\usepackage{amssymb,amsmath,array,bm}
\usepackage{booktabs}

\usepackage{tikz}
\usetikzlibrary{arrows}

\usepackage{pgfplots}
\pgfplotsset{compat=1.15}
\usetikzlibrary{arrows}

\usepackage[a4paper, left=3cm, right=2.5cm, top=3cm, bottom=3cm]{geometry}

\usepackage{authblk}
\usepackage{natbib}
\usepackage{doi}

\usepackage[utf8]{inputenc}
\usepackage[T1]{fontenc}
\usepackage[english]{babel}
\usepackage{csquotes}

\title{16-bit Precision of Convolutional Neural Networks on Microcontroller Units for 8-bit Costs}

\author[1]{Rui Liu}
\author[1]{Benjamin Paaßen}
\affil[1]{Faculty of Technology, Bielefeld University}

\date{preprint as provided by the authors}

\begin{document}

\maketitle

\pagestyle{myheadings}
\markright{preprint as provided by the authors}

\begin{abstract}
To deploy deep neural networks on edge hardware, highly efficient inference schemes are necessary that retain high accuracy. This work presents W16A16, a high precision (16-bit), fast speed, low energy quantization method. On a widely applied microcontroller architecture Armv7E-M, our proposed approach achieves faster speed and lower energy consumption on layer- and model-level compared to alternative quantization schemes.
We analyze the architecture of Armv7E-M, explain the underlying principles behind the performance advantages of 16-bit approaches, and evaluate the empiric quantization errors for regression and classification tasks, as well as empiric time- and energy consumption in MCU deployment. We observe ca.\ 10 times lower quantization errors compared to 8-bit quantization schemes while achieving similar or better inference times and energy consumptions.

\end{abstract}

\section{Introduction}

Deep Neural Networks (DNNs) have been widely applied in various domains, such as image recognition \citep{he2016deep}, speech detection \citep{nassif2019speech} and signal regression \citep{qiu2014ensemble}. However, executing DNNs (also called inference) often requires substantial computational resources and energy \citep{zaidi2022survey}. This resource demand often makes it infeasible to deploy DNNs on resource-limited platforms, such as Microcontroller Units (MCUs) and Microprocessing Units (MPUs). MCUs are the mainstream choice for the deployment in industrial and Internet of Things (IoT) devices \citep{novac2021quantization}. Therefore, making DNNs executable on MCUs would unlock substantial capabilities for such edge devices.

Typical MCUs have very limited memory, often less than 1 MB, and low clock speeds (often 40 to 500 Mhz) \citep{deutel2023energy}. However, industrial and IoT scenarios often demand very small delays (real-time reaction), particularly in safety-critical applications, such as Health Monitoring \citep{alshehri2020comprehensive}, industrial IoT, predictive maintenance and autonomous vehicles \citep{yurtsever2020survey}. As such, DNNs in such scenarios not only need to be small (in terms of memory) but also fast (in terms of cycles required for executing an entire DNN inference).

To overcome the challenge of limited resources on edge devices, the research field of TinyML emerged, focusing on the deployment of complex DNNs models by means of novel algorithms, hardware-software co-designs, neural network pruning \citep{liberis2023differentiable} and quantization \citep{rusci2020memory}. Among these methods, quantization has become the most widely utilized approach \citep{nagel2021white}. Quantization refers to the rounding of 32-bit floating point numbers (which typically represents weights and activations in DNNs) to lower-resolution integers (e.g. 8-bit, 16-bit). This does not only reduce the memory footprint of DNNs but also enables the use of integer arithmetic components on the MCU, which are typically much faster and more energy-efficient compared to floating point units. Consequently, 8-bit integer (INT8) quantization has been been widely used and deeply optimized by various embedded AI frameworks and libraries, such as Tensorflow-Micro-Light \citep{david2021tensorflow}, X-CUBE-AI \citep{falbo2019analyzing}, and CMSIS-NN \citep{lai2018cmsis}. Indeed, INT8 quantization has become a standard approach for MCUs \citep{abushahla2025neural}. However, the loss of resolution due to INT8 quantization is so severe that the accuracy of DNNs may be affected, especially for regression tasks (as we show in our experiments). As such, higher resolution via 16-bit quantization would be desirable---but may risk a higher energy- and memory footprint. Fortunately, in this paper, we show that the instruction set of the widely used ARMv7E-M architecture enables us to run INT16 quantized neural nets with almost the same time- and energy efficiency compared to INT8 quantized nets while achieving almost the same accuracy as the full FLOAT32 network. In other words: we get 16-bit resolution for the cost of 8-bit resolution.

In summary, our contributions are: (1) We analyze the ARMv7E-M architecture and its assembly instruction set to derive a novel approach for INT16 quantized DNNs on MCUs. (2) We implement this approach in software for the example of temporal convolutional networks (TCNs) and provide this implementation as an open source package. (3) We empirically demonstrate superior performance of INT16 TCNs compared to INT8 TCNs in both classification and regression tasks, on par with the full FLOAT32 version of the model. (4) We empirically demonstrate that the number of computations, inference time, and energy consumption of our proposed INT16 implementation is comparable to INT8 quantization and outperforms a mixed quantization scheme (8-bit weights, 16-bit activations) as well as reference quantization schemes implemented as part of the Cube.AI library.

The implementation of our approach is available at \url{https://gitlab.com/fastconv1d/W16A16}.

\section{Background and Motivation}\label{}
\subsection{Integer Quantization Foundations}
Unlike general-purpose x86 CPUs in PCs or Cortex-A cores in mobile devices, MCU cores such as Cortex-M work under extremely constrained memory and power conditions, and are not suitable for extensive floating point calculations that consume lots of energy. Therefore, integer operations are generally preferred on MCUs, which suggests quantization approaches for DNNs \citep{abushahla2025neural}.
The quantization of a floating point number $X_{float}$ into an integer representation can be computed via the following equation:
\begin{equation}
\label{eq:quantization}
X_\text{int} = \text{clip}\left( \text{round}\left( \frac{X_\text{float}}{S} + Z \right), Q_{min}, Q_{max} \right),
\end{equation}
where $X_\text{float}$ is the floating-point input, $X_\text{int}$ is the corresponding integer-quantized result, $S$ is the conversion rate of floating points to integer numbers, meaning how many floating point units map to each step in the integer scale, $Z$ is an integer value that corresponds to the mid-point of the float scale, and $Q_{min}$ and $Q_{max}$ are the minimum and maximum quantization values.
For example, assume that we expect floating point numbers in the range $[-1.0, +3.0]$ and we want to quantize these numbers into 8 bits. Then, we should set $S = \frac{4}{2^8} = 0.015625$ to convert the float range into the integer range, $Z = -\frac{1}{S} = -64$ to account for the expected mid-point (namely +1, the mid-point between -1 and +3) of the float scale, as well as $Q_{min} = -2^{8-1} = -128$ and $Q_{max} = 2^{8-1}-1 = 127$ to account for the end points of the scale. With these parameters, $-1.0$ maps exactly to $-128$, $+3.0$ maps to $128$ (and gets clipped to $127$) and $1$ (the mid point) maps to $0$. 
The example also shows that the parameters of the quantization should ideally be fitted to a distribution of expected weights and activations to avoid unnecessary precision losses due to rounding or clipping. In our example, any numbers above $+3.0$, such as $+3.5$ or $+12.0$, will be clipped to $127$.

\begin{figure}
\includegraphics[width=6cm]{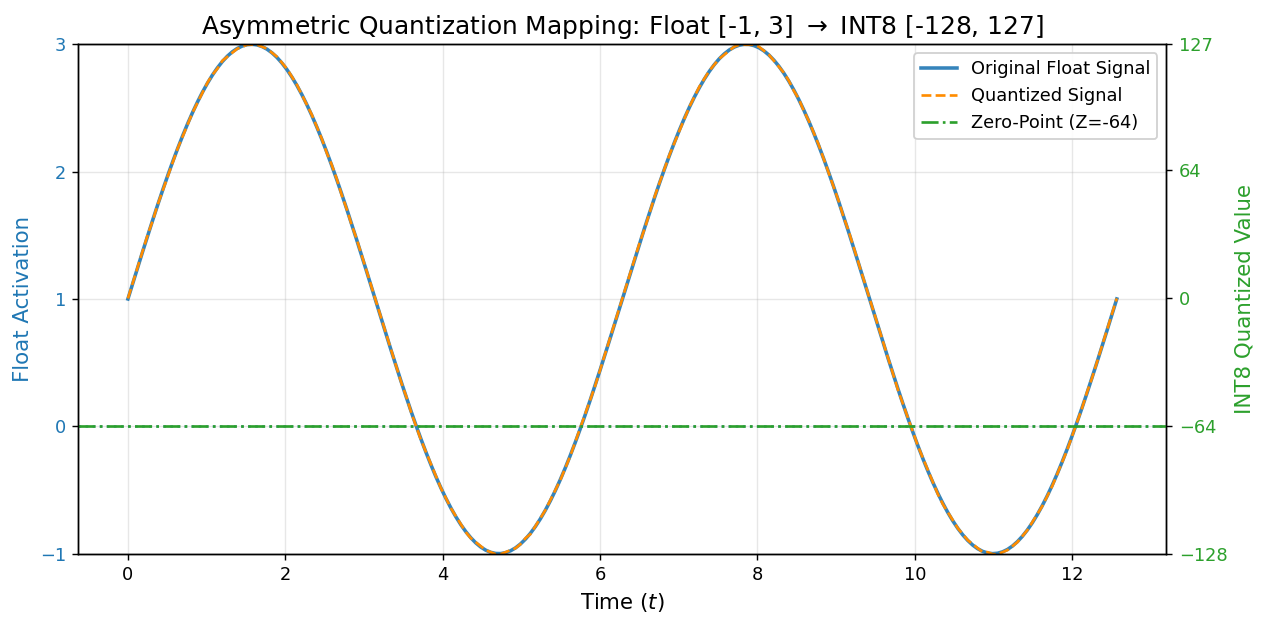} 
\centering
\caption{The asymmetric quantization of activation values.} 
\label{fig:asymmetric_quantization}
\end{figure} 

\begin{figure}
\includegraphics[width=6cm]{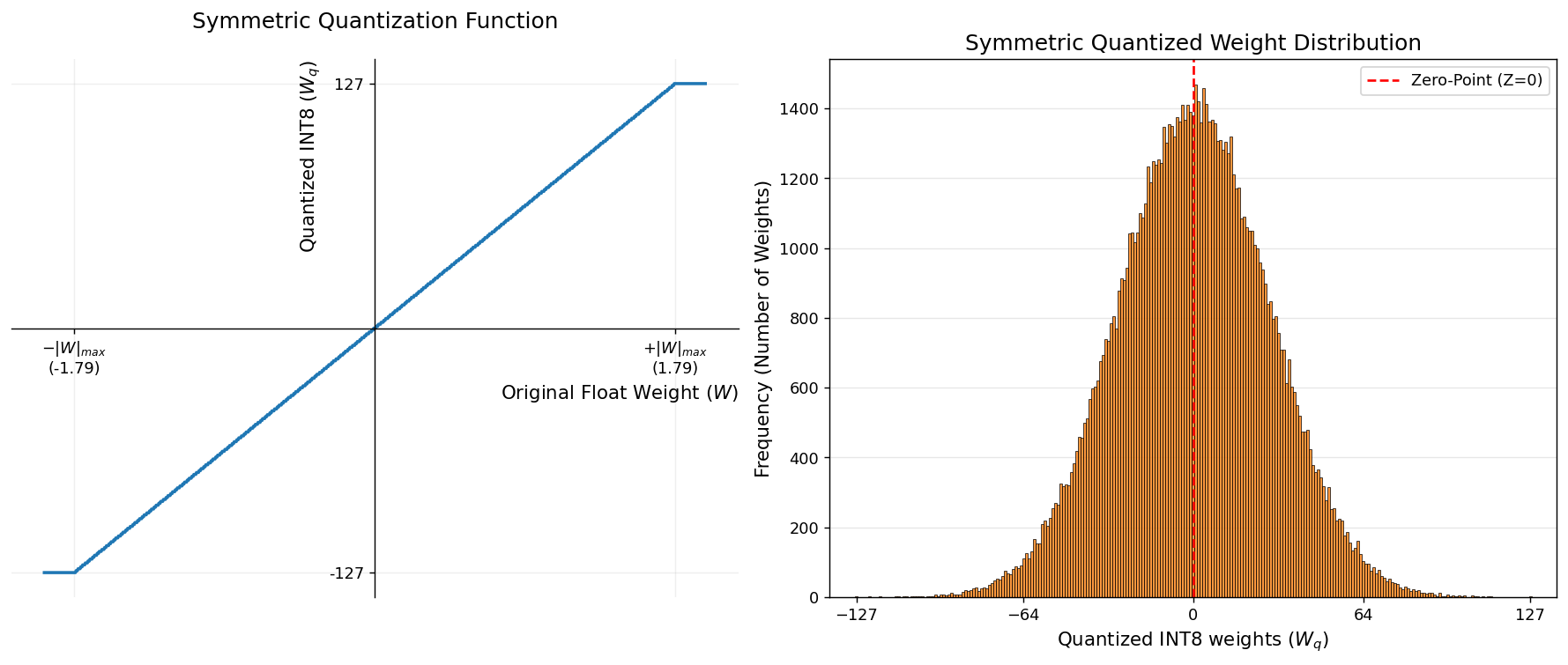} 
\centering
\caption{The symmetric quantization of weight values.} 
\label{fig:symmetric_quantization}
\end{figure} 

The quantization reduces the memory footprint of the represented number and accelerates computations because integer operations are generally faster than floating point operations. When dealing with deep neural nets (DNNs), quantization is applied to both weights and activations to speed up the execution of the DNN. 
Because different resolutions may be applied to weights and activations, we will adopt a notation that makes the resolution explicit for both weights and activations, namely the letter W, followed by the resolutions for the weights of a DNN, followed by the letter A, followed by the resolution for the activations of a DNN. Our proposed quantization scheme in this work is W16A16, meaning 16-bit quantization for both weights and activations.

\subsubsection{Static v.s. Dynamic Quantization}

The aforementioned quantization formula~\eqref{eq:quantization} is also called \emph{static} quantization, which sets the quantization parameters $S$, $Z$, $Q_{min}$ and $Q_{max}$ statically before the DNN is executed. By contrast, it is also possible to adjust these parameters dynamically to the current activations during DNN execution to better match the incoming data \citep{zhao2024atom}.
However, dynamic quantization requires substantial computation overhead to select the correct parameters, which is infeasible on MCUs. Therefore, TinyML, including our approach in this paper, employs static quantization to avoid computationally intensive floating-point operations.
%

\subsubsection{Symmetric vs. Asymmetric Quantization}

Quantization with $Z = 0$ is called symmetric, whereas quantization with $Z \neq 0$ is called asymmetric \citep{gholami2022survey}. The weight distribution of convolutional neural networks is typically symmetric around zero, such that symmetric quantization is viable \citep{lai2018cmsis} (refer to Fig.~\ref{fig:symmetric_quantization}). However, the distribution of activations may well be asymmetric (refer to Fig.~\ref{fig:asymmetric_quantization}), suggesting asymmetric quantization. However, asymmetric quantization comes at a price: a convolutional layer involves thousands of weight-activation multiplications of the form $w_1 \times x_1$. If the activations are quantized asymmetrically, these operations always require the form $w_1 \times (x_1-Z)$, meaning $Z$ has to be subtracted before the multiplication can be executed. This leads to an increase in execution cycles and register usage. For 8-bit quantization, this overhead unavoidable without major quantization errors. However, for our proposed 16-bit quantization, the value range is so much larger (-32768 to +32767 instead of -128 to 127) that the entire activation range can fit inside without major errors (as we will show in the experiments). Hence, our proposed W16A16 scheme will use symmetric quantization for both weights and activations.

\subsubsection{Per-layer vs. per-channel quantization}

Quantization can also be distinguished into per-layer and per-channel \citep{abushahla2025neural}. Per-layer quantization employs the same quantization parameters for each floating point number in an entire layer of a DNN, whereas per-channel quantization uses different parameters per output of weights, thus accounting for inter-channel variation. In practice, it is a standard method for convolutional neural network due to channel-wise variation \citep{abushahla2025neural}. Accordingly, this paper will employ per-layer for activations and per-channel quantization for weights.

\subsection{Multiply-and-accumulate operations over quantized data}
\label{sec:mac}

The core of a convolutional layer is the convolution of weights and inputs. Assume an input matrix $X$ of shape $(T, IC)$, where $T$ is the number of data points and $IC$ is the number of input channels; and a weight tensor $W$ of shape $(OC, K, IC)$, where $K$ is the size of the convolutional kernel and OC is the number of output channels. Then, the convolutional operation (leaving bias terms aside for simplicity) is given via the equation:

$$y_{t, c} = \sum_{i=1}^{IC} \sum_{k=1}^K X_{t, i} \cdot W_{c, k, i}$$

This operation can be effectively implemented via multi\-ply-and-accumulate (MAC) operations in MCUs. Each MAC operation multiplies one input with one weight and adds the result to an accumulator register. Because MAC operations are so frequent in the inference computations of DNNs, the speed of MAC operations determines the overall execution speed of DNNs on MCUs \citep{sze2017efficient}.

When implementing MAC operations in quantized form, the respective scaling factors have to be taken into account. Assume a scaling factor $S_{in}$ for the inputs of a convolutional layer, a scaling factor $S_{weight, c}$ for the weights in output channel $c$ of the layer, and a scaling factor $S_{out}$ for the output. Then, a single quantized MAC operation corresponds to

\begin{align*}
&Accumulator'_{int} \times (S_{in} \times S_{weight, c}) \\
= &Accumulator_{int} \times  (S_{in} \cdot S_{weight, c})\\
&+(X_{int} \times S_{in}) \cdot (S_{weight, c} \cdot W_c),
\end{align*}
where $Accumulator_{int}$ denotes the accumulated value before the MAC operation, and $Accumulator'_{int}$ denotes the accumulated value after the MAC operation.
In other words, we perform an integer multiplication of the current quantized input with the current quantized weight and add this to the current content of the accumulation register, yielding a new accumulator value which is quantized with the scaling factor $S_{in} \times S_{weight, c}$. After all MAC operations are computed, the accumulator corresponds to $y_{t, c}$, i.e.\ the output value for the $c$th output channel.

\subsubsection{Re-quantization}
\label{sec:Re-quantization}

After the MAC operations for one layer of a DNN are completed, the content of the accumulation register needs to be re-quantized to be usable as an input for the next layer. This has two reasons: First, to ensure that no overflows occur, the accumulator for the MAC operations needs to have a substantially higher bit width (e.g. 64 bits) compared to the quantized weights and inputs. This needs to be reduced back to the activation bit width. Second, the scaling factor for the accumulator is $S_{in} \times S_{weight, c}$, which needs to be converted to the scaling factor $S_{out}$ for activations of the next layer.

More formally, we wish to obtain an integer-quantized output as follows:
$$Output_{int} \times S_{out} = Accumulator \times (S_{in} \times S_{weight, c})$$

Accordingly, the integer quantized output result is:
\begin{equation}
\label{eq:requantization}Output_{int} = Accumulator \times \left( \frac{S_{in} \times S_{weight, c}}{S_{out}} \right)\end{equation}

The requantization scaling factor $S_{requantization,c}$ is:
\begin{equation}
\label{eq:requantization_factor}S_{requantization, c} = \frac{S_{in} \times S_{weight, c}}{S_{out}}\end{equation}

In summary: To re-quantize the content of the accumulation register, we first need to multiply the accumulator with $S_{requantization, c}$ and then need to round the result back to 8 bits or 16 bits. For W8A8 quantization, this multiplication and rounding is implemented in the \texttt{arm\_nn\_requantize} function. For W8A16 quantization, the \texttt{arm\_nn\_requantize\_s64} can be used. For our proposed W16A16 quantization, no suitable re-quantization function is available and we introduce a custom version in Sec.~\ref{sec:requantization_fixed_point}



\subsubsection{Fixed-point approximation of float scaling}
\label{sec:fixed_point}
The scaling factor $S_{requantization, c}$ is actually a floating-point number, which would make Equation~\eqref{eq:requantization} a floating point multiplication. To avoid such a float operation, we wish to approximate the multiplication in Equation~\eqref{eq:requantization} with integer operations.

In the early version of CMSIS-NN \citep{lai2018cmsis}, a Q-format fixed-point quantization (e.g. Q15/Q7) is performed to approximate the scaling factor multiplication, e.g.\ via the \texttt{arm\_convolve\_HWC\_q15} function.
This scheme approximates the multiplication with $S$ by an $n$-bit shift. Accordingly, only powers of two are available as scaling factors. However, actual scaling factors are rarely exact powers of two, resulting in large quantization errors. For example, a scaling factor of $0.1$ would have to be approximated by the nearest power of 2, namely $2^{-3} = 0.125$, resulting in an error of 25\%.

Therefore, more accurate approximation schemes have been developed, which instead decompose the scaling factor $S$ as $S \approx M_{Q31} \times 2^{-31-n}$, where $M_{Q31}$ is a 32-bit integer, such that more values in the range $[0, 1]$ can be represented. \citep{jacob2018quantization}. For example, a multiplication with $S = 0.1$ can be approximated as first multiplying with $M_{Q31} = 1717986918$ and then shifting by $31+3 = 34$ bits, resulting in $1717986918 \times 2^{-34} \approx 0.1$ with an error less than $10^{-9}$.

To guarantee high precision, our W16A16 adopts the Q31 fixed-point approximation method (a multiplier combined with right shifts) of \citep{jacob2018quantization} to approximate the multiplication with scaling factors.

\subsection{The Cortex-M processor instruction set}

\subsubsection{ARMv7E-M Dual MAC operations}
The Cortex-M processor applies sub-word parallelism to maximize throughput. According to \citep{lai2018cmsis}, the ARMv7E-M architecture has the capability to divide each 32-bit register into two 16-bit sub-words which enables dual multiplication and accumulation (MAC) instructions (e.g. SMLAD and SMLALD). SMLAD executes double INT16 x INT16 MAC within a single clock cycle. However, SMLAD doesn't support four INT8 x INT8 MAC operations. Accordingly, INT8 and INT16 MAC operations are similarly fast on the ARMv7E-M. Indeed, INT8 operations introduce some overhead because operands first need to be unpacked and preprocessed before SMLAD instructions can be executed, whereas INT16 operands have the correct width right away.

Importantly, most operations involved in executing a convolutional neural network (or any DNN) are MAC operations. Thus, INT8 and INT16 quantized DNNs are similarly fast.


Taking the INT8 convolution layer operator \texttt{arm\_convolve\_s8} from the CMSIS-NN library as an example, sign extension instruction and re-ordering instruction are required for data preprocessing.
In the \texttt{arm\_convolve\_s8} function, the \emph{Sign extend two Bytes with Add} instruction (SXTAB16) extends 8-bit weight data into a 16-bit format and performs a ZeroPoint addition for asymmetric quantization. Finally, the re-ordering instructions PKHTB and PKHBT are used to ensure the correct order of the output data.

By contrast, loading INT16 data does not require such data pre- or postprocessing instructions. This results in a speed advantage of INT16 compared to INT8 processing. The detailed comparison of INT8 and INT16 precessing on the instruction level is provided in Section~\ref{Instruction_Analysis}.

\subsubsection{Single-issue vs. Dual-issue}
The M4 and M7 Micro-Architecture utilize the same ARMv7E-M DSP Instruction Set. However, they differ in the execution flows. The Cortex-M4 processor allows only a single instruction to be executed per clock cycle following the in-order strictly.

By contrast, Cortex-M7 employs a 6-stage superscalar pipeline architecture with dual-issue capability \citep{lorenser2016dsp}. The Dual-issue capability allows the processor to execute two instructions within the same clock cycle, thus increasing inference speed. 
However, Instruction-level parallelism (ILP) has strict requirements: The dual-issue pair must not have data dependencies such as Read After Write (RAW). 
Our paper evaluates different quantization methods with both the M4 and the M7 Micro-Architecture to ensure the advantages of W16A16 quantization method across two architectures.

\subsection{Limitations of Existing Inference Frameworks}
Currently, the deployment of quantized neural networks is dominated by two standard end-to-end inference engines. TFLM is an open-source engine developed by Google \citep{david2021tensorflow} supporting W8A8 and Mixed-precision W8A16 quantization approaches. However, TFLM triggers an error when implementing a TCN, because it just supports SAME/VALID padding. Therefore, TFLM cannot be applied for the quantization and implementation of our TCN on MCUs.

X-CUBE-AI, developed by STMicroelectronics also exhibits a similar issue: when implementing TCNs, the computation units fall back to floating-point operation, resulting in a performance drop and energy waste. Furthermore, X-CUBE-AI supports only INT8 quantization, lacking the INT16 quantization format.
Thereby, TFLM and X-CUBE-AI are infeasible for the effective implementation of quantized TCNs. Still, our paper utilizes X-CUBE-AI as one of the baselines in the experiments.

\section{Methods}

This paper proposes a novel W16A16 quantization scheme for ARMv7E-M MCUs for DNN inference. We begin with temporal convolutional networks (TCNs) as a reference neural network architecture to be quantized (Sec.~\ref{sec:TCN_architecture}) and then introduce the W16A16 quantization scheme, discussing in detail the assembly instructions used (Sec.~\ref{Instruction_Analysis}), the memory layout (Sec.~\ref{sec:memory_layout}), and the requantization method (Sec.~\ref{sec:requantization_fixed_point} and Sec.~\ref{sec:two_stage_saturation}).

\subsection{Temporal Convolutional Network Quantization}
\label{sec:TCN_architecture}

\begin{figure}
\includegraphics[width=6cm]{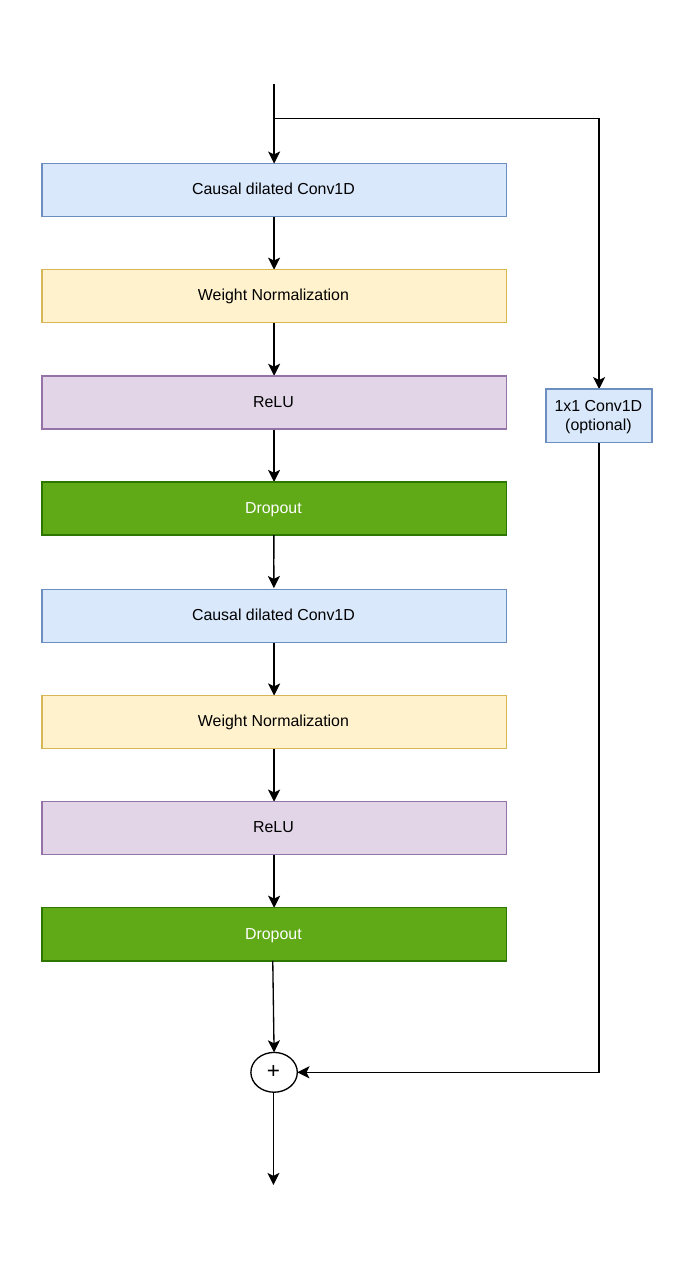} 
\centering
\caption{The structure of a TCN block.} 
\label{fig:TCN_block}
\end{figure} 

Temporal Convolutional Networks (TCNs) are used for classification and regression on time series data \citep{lin2021temporal}. Unlike Recurrent Neural Network (RNN) or Long short term Memory (LSTM), which require step-by-step processing with sigmoid nonlinearities due to recurrent data dependencies, TCNs are highly parallelizable, similar to 2D-CNNs for images. The core of a TCN is a dilated 1D convolution, as defined in the following equation.
\begin{equation}
\label{eq:dilated}
y_{c, t} =  \sum_{k=0}^{K-1}\sum_{i=0}^{IC-1} W_{c, k, i} \cdot X_{i, t - D \cdot k} + B_c,
\end{equation}
where $c$ indicates an output channel, $i$ indicates an input channel, $k$ runs over the kernel size $K$, $D$ is the dilation across time, $W$ indicates a weight, $B$ a bias, and $X$ the input data (from the previous layer).

The TCN structure applied in our paper is summarized in Fig.~\ref{fig:TCN_block}: the TCN block consists of two causal dilated 1D-convolutional layers, interleaved by weight normalization, a rectified linear unit (ReLU) and a dropout layer, following \cite{lin2021temporal}. Optionally, a 1D-convolutional layer in the residual connection can be added to adjust the feature map size. 
A full TCN consists of a stack of TCN blocks followed by a single 1D-convolutional layer to compute the overall output of the TCN.

In the remainder of this section, we introduce W16A16 quantization for TCNs, and design quantized TCN operators for MCUs. 



\subsection{Assembly Instruction Analysis}
\label{Instruction_Analysis}
A convolutional neural network of realistic size requires millions of MAC operatations. Therefore, the SMLAD instruction (dual-MAC execution instruction) effectively determines the execution speed of convolutional layers.
In this section, we analyze the micro-architecture of Cortex-M CPUs and the instruction efficiency of W8A8, W8A16 and our proposed W16A16 quantization method in implementing SMLAD instructions. To analyze the differences between the quantization schemes, we use the running example of performing MAC on four inputs ($X_0$, $X_1$, $X_2$, $X_3$) with four weights ($W_0$, $W_1$, $W_2$, $W_3$).


\begin{figure}
\includegraphics[width=8cm]{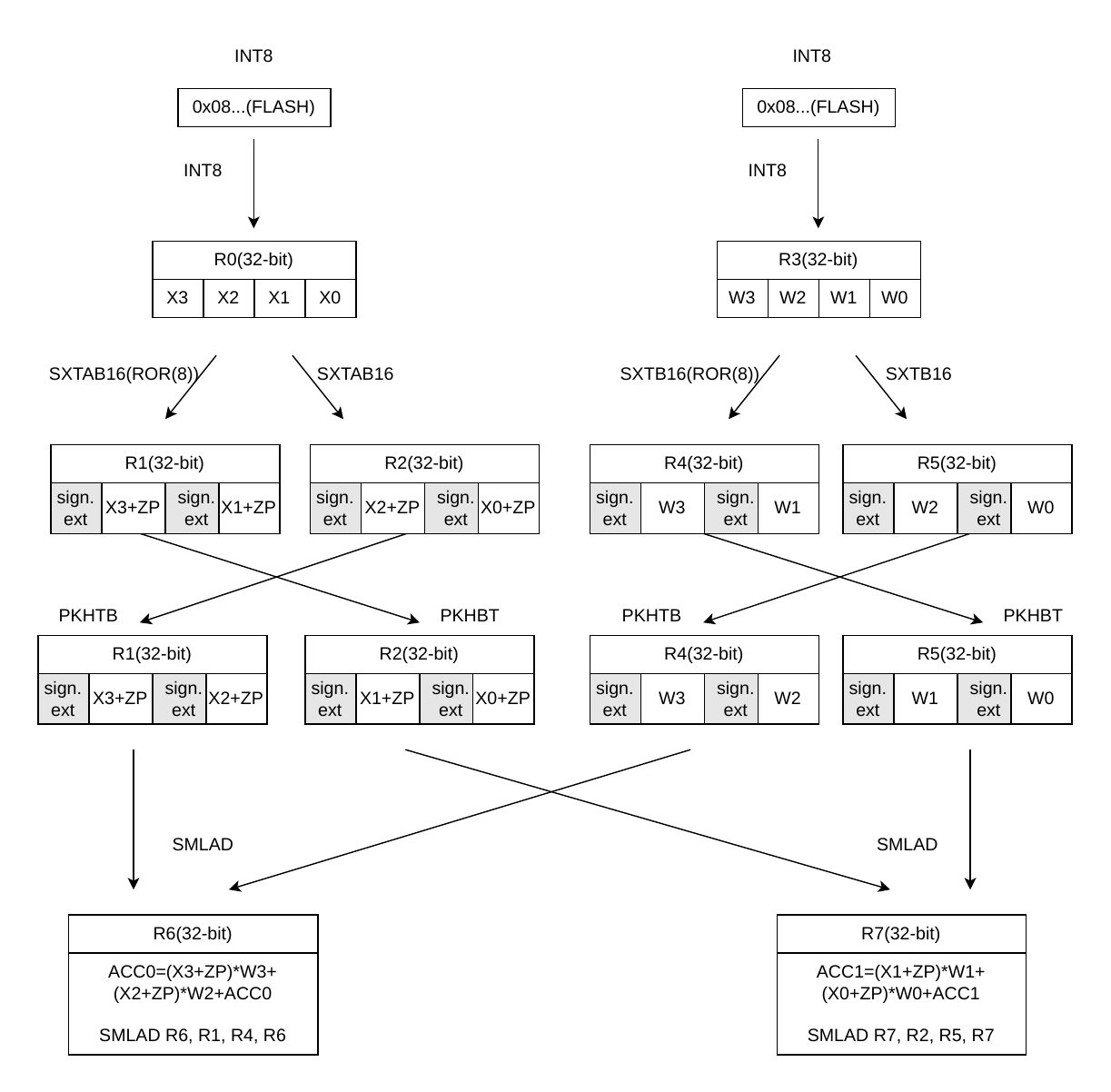} 
\centering
\caption{Instructions needed to perform MAC operations for four inputs and four weights in W8A8 quantization} 
\label{fig:W8A8}
\end{figure} 


The W8A8 scheme applies asymmetric quantization for activations and symmetric quantization for weights. Accordingly, the SXTAB16 instruction is needed to perform sign extension and adding the offset $Z$ within one Instruction. As Fig.~\ref{fig:W8A8} shows, W8A8 quantization requires two LDR instructions to load four inputs and four weights, followed by eight instructions for data preprocessing, namely two SXTAB16 for input data, two SXTB16 for weight data and four PKHBT / PKHTB for re-ordering. Finally, two MAC instructions (SMLAD) are executed. The total amount is 12 instructions.

The W8A16 scheme employs symmetric quantization for both activations and weights, requiring three LDR to load data, namely one LDR for INT8 weights and two LDR for INT16 inputs. Then, four data preprocessing operations instruction are needed, followed by two MAC operations. The total amount is 9 instructions.

Our W16A16 scheme utilizes four LDR to load input data and weights and two MAC instructions. The total amount is 6 instructions. Note that the multiplication of two 16-bit numbers and performing hundreds of accumulations may exceed a 32-bit integer, resulting in a potential overflow problem. Therefore, we use the SMLALD instruction to store the result in a 64-bit register, instead. Note that the speed of SMLALD and SMLAD are the same.

Compared to W8A8, our proposed W16A16 does not require any sign extension and re-ordering instructions, reducing the total number of instructions to 6, a saving of 50\%. Furthermore, data dependencies exist among the Load (LDR), Sign extension (SXTAB16 / SXTB16), Re-ordering (PKHBT / PKHTB) and Dual-MAC (SMLAD), such that these instructions must be executed sequentially, potentially causing pipeline stalls and preventing acceleration via dual-issue in the M7 architecture. By eliminating the data preprocessing instructions (such as SXTB16), the register pressure can also be decreased. Our proposed W16A16 achieves zero-overhead alignment; the data fetched from FLASH is fed directly into dual-MAC (SMLAD / SMLALD) instruction, thereby minimizing redundant instructions.

However, compared to W8A8, W16A16 requires two additional LDR instructions to load the same amount of input data and weights. 
Therefore, further experiments are needed to verify which of the W8A8, W8A16, or W16A16 quantization approaches yields actual faster inference speeds and lower energy consumption.
\begin{figure}
\includegraphics[width=8cm]{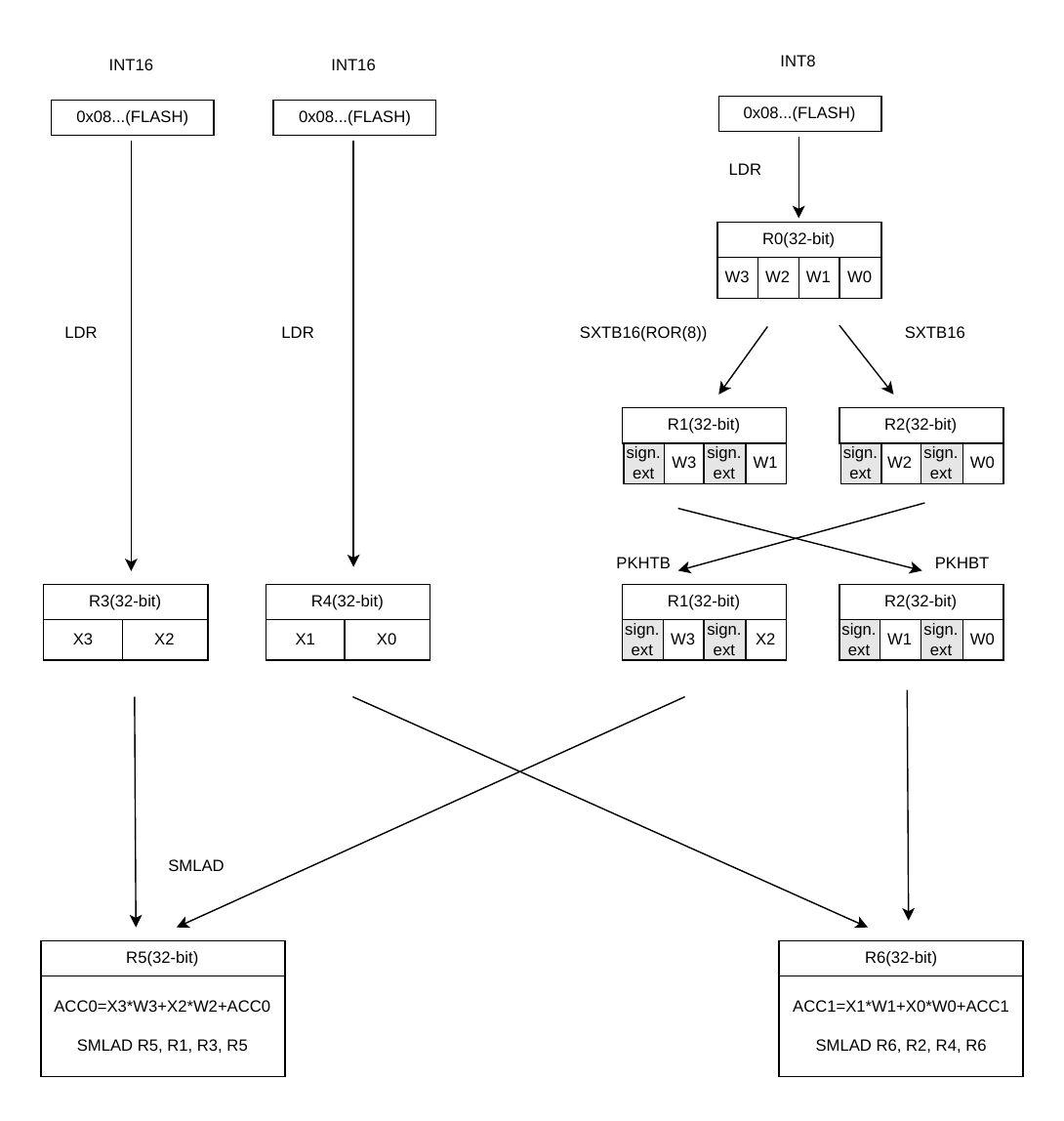} 
\centering
\caption{Instructions needed to perform MAC operations for four inputs and four weights in W8A16 quantization} 
\label{fig:W8A16}
\end{figure}

\begin{figure}
\includegraphics[width=8cm]{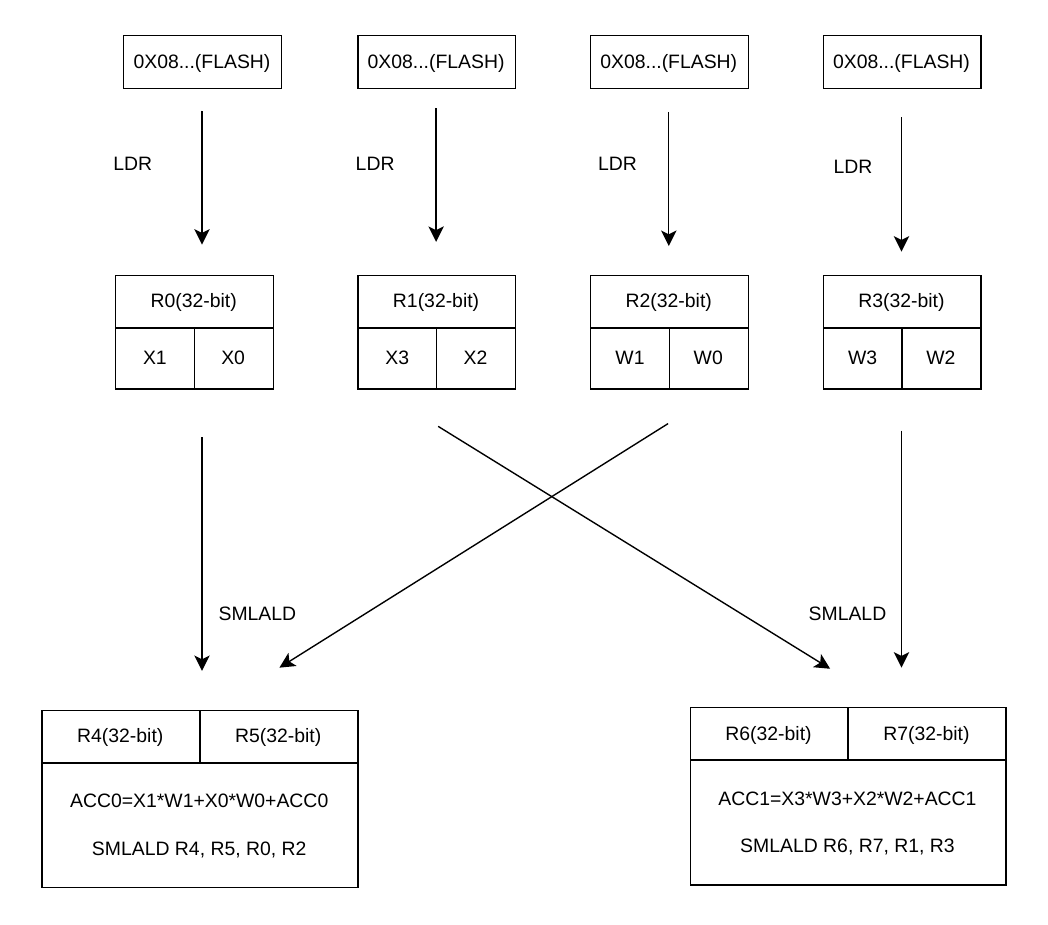} 
\centering
\caption{Instructions needed to perform MAC operations for four inputs and four weights in W16A16 quantization} 
\label{fig:W16A16}
\end{figure}

\subsection{Memory Layout and Convolutional operator design}
\label{sec:memory_layout}
Before discussing the design and implementation of convolutional operators on MCUs in detail, it is necessary to first explain the memory layout of input and weight data.

\begin{figure}
\includegraphics[width=8cm]{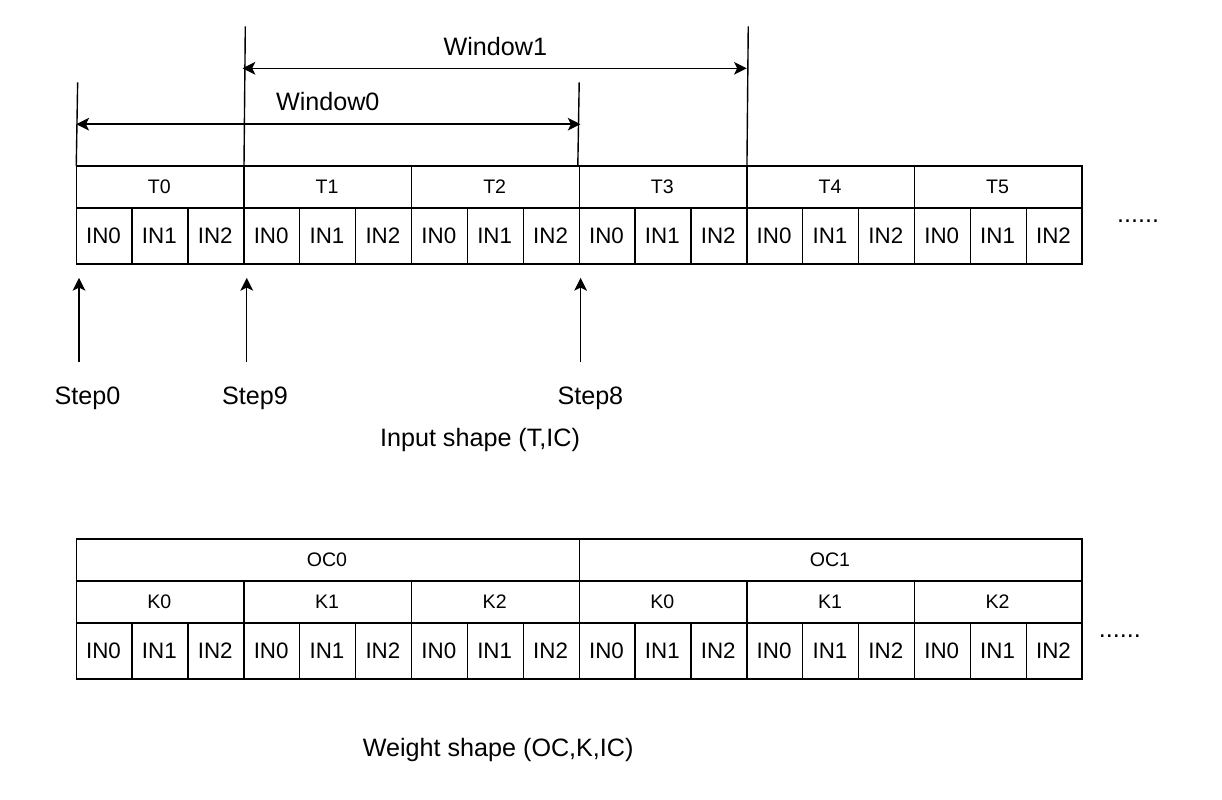} 
\centering
\caption{The memory layout} 
\label{fig:memory}
\end{figure}

Fig.~\ref{fig:memory} shows an example of the implementation of a 1D convolutional layer. The input data has a shape of $(T, IC)$, where $T$ is the number of time steps and $IC$ the number of input channels, and each entry in the memory represents the data of one input channel at one time step. In this example, the number of input channels is three, meaning each time step T contains three elements IN0, IN1, and IN2.

The shape of the weight tensor is $(OC, K, IC)$, where OC denotes the number of output channels and K denotes the size of kernel. As Fig.~\ref{fig:memory} illustrates, the lowest-level data in memory consists of input channels, grouped into packages for each kernel window (K0, K1, K2), which in turn are grouped into packages for each output channel (OC0, OC1, ...). 

Accordingly, the memory layouts of inputs and weights do not match. When executing the convolution operation, the 1DCNN's receptive field (with kernel size K = 3) needs to slide as a window across the input data. As Fig.~\ref{fig:memory} shows, after the CPU finishes processing all elements within the receptive field of the first Window0 at Step 8, it must jump back to the location to Step 9, which is the start point of the next window. This jump operation is a discontinuous memory access, resulting in longer access times.

For 2D convolution operations, the Im2col (image to column) method is proposed to address this problem by flattening and rearranging the input data within a receptive field \citep{chellapilla2006high}. This method transforms the complicated multi-channels sliding window into a one-dimensional General Matrix Multiplication (GEMM).

\begin{figure}
\includegraphics[width=8cm]{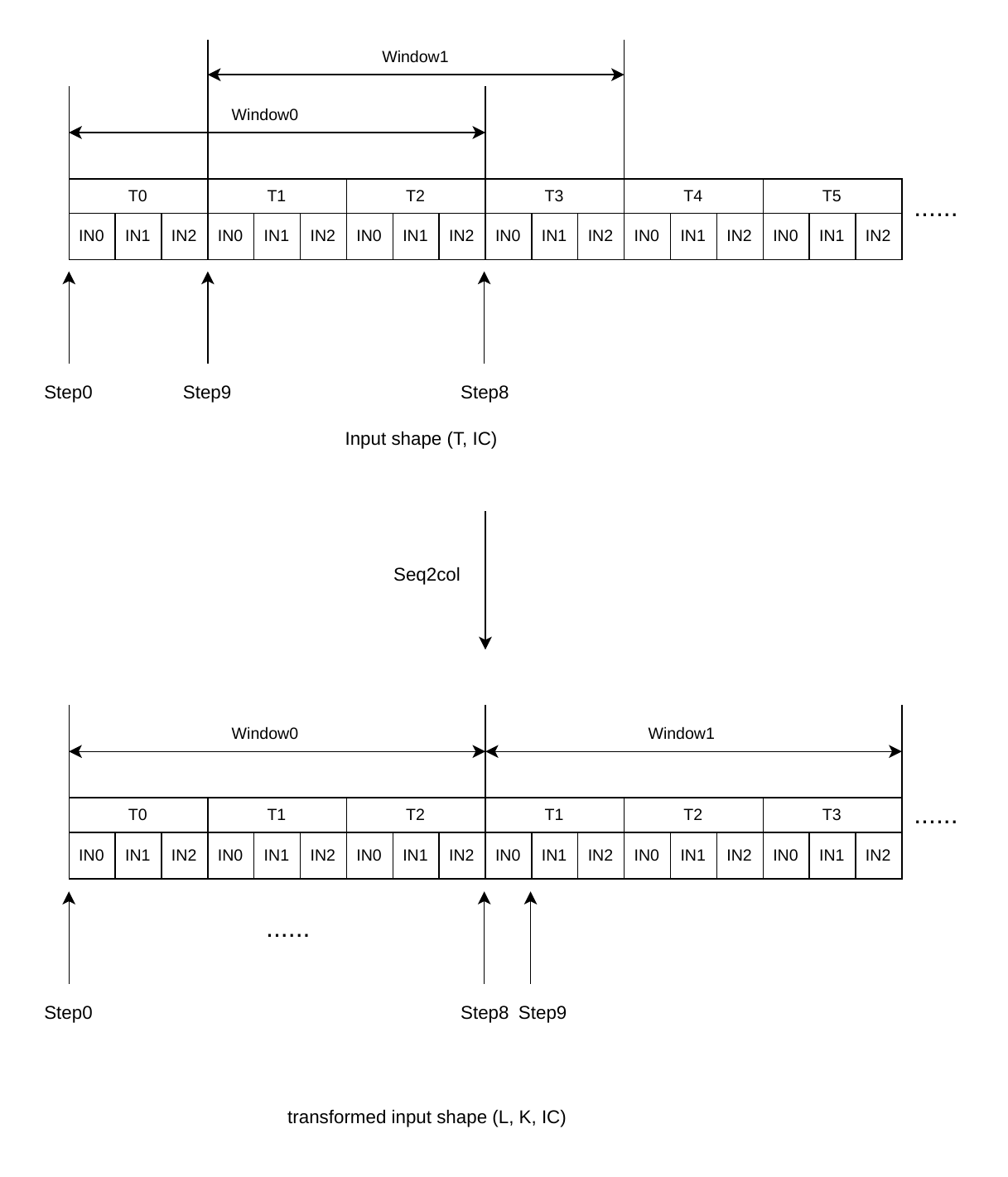} 
\centering
\caption{A comparison of discontinuous array access in the original memory layout (top) versus the sequential array access after Seq2col (bottom).} 
\label{fig:principles_Seq2col}
\end{figure}

For our time series data, we also need a transformation like Im2col. For the 1D case, we name the transformation Seq2col (sequence to column). Fig.~\ref{fig:principles_Seq2col} demonstrates the principle of Seq2Col. The sliding window process is decomposed and flattened into a series of windows, where $L$ denotes the number of windows.



After the processing of Seq2col, the input data shape is converted into the shape $(L, IC, K)$, while the shape of the weights is $(OC, IC, K)$. 
As Fig.~\ref{fig:principles_Seq2col} shows, the windows are put into sequential order after Seq2col, thereby simplifying memory access.



\begin{figure}
\includegraphics[width=8cm]{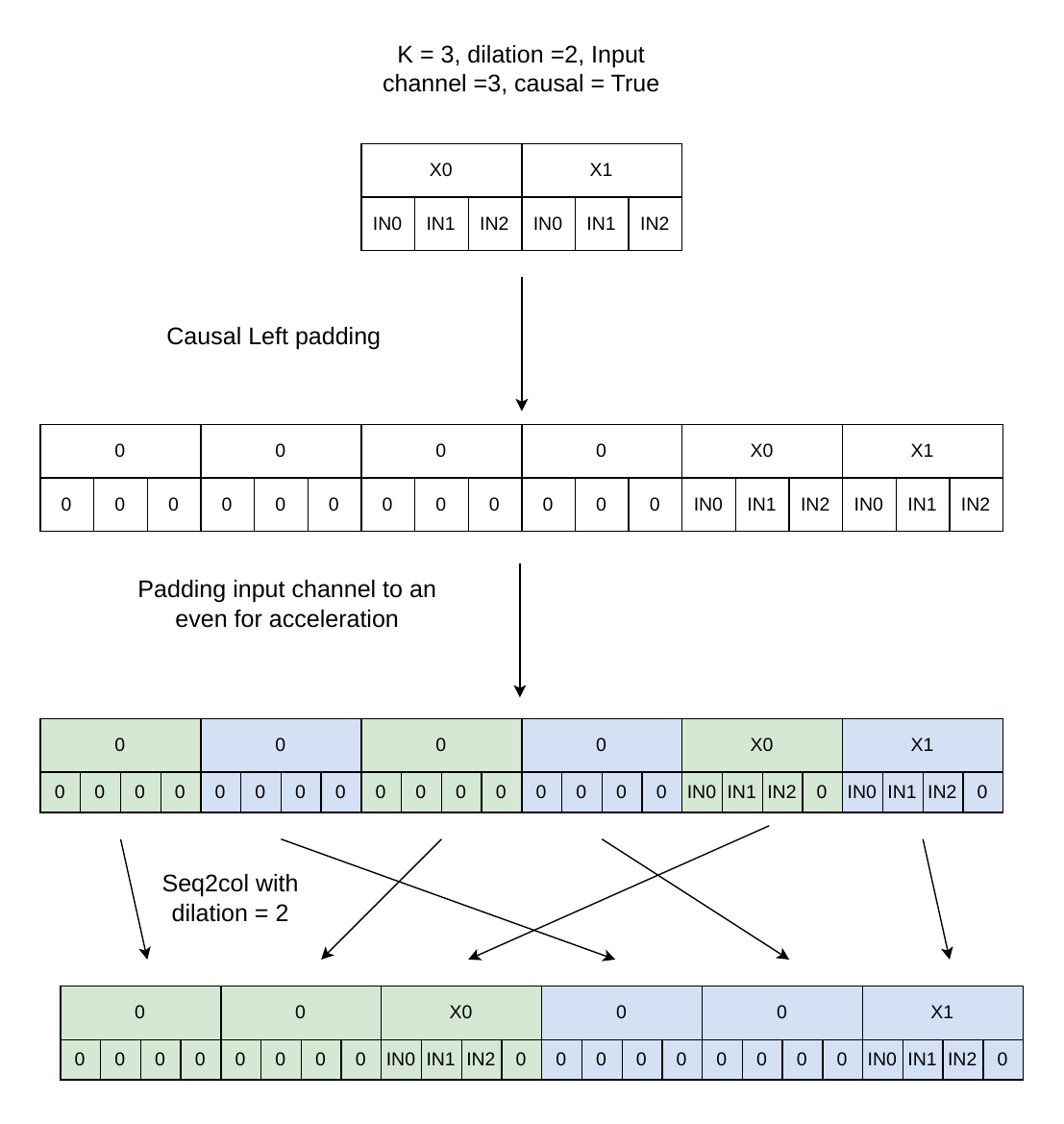} 
\centering
\caption{An example application of Seq2Col, including causal left padding and dilation.} 
\label{fig:seq2col}
\end{figure}

Note that, in our work, we designed a Seq2col method for a 1D-dilated convolutional layer, which requires asymmetric left padding and dilated computing.
In Fig.~\ref{fig:seq2col}, we use a causal convolutional layer with kernel size $K=3$, dilation $d=2$, and 3 input channels as example. First, causal left padding of size $(K - 1) \times d$ is applied to ensure the causal relationship of data.
Second, the input channels (e.g. IN0, IN1, IN2) are padded to an even number (e.g. 3 to 4) to ensure memory addresses are aligned to multiples of four.
Thereby, dual-word load instructions such as LDRD can be utilized to accelerate data loading; this instruction requires the address to be aligned to a multiple of 4 and its execution speed is twice that of LDR. 

Third, following the padding phase, the receptive field of a causal dilated convolutional layer is constructed by fetching the discontinuous dilated input data from original input data, resulting in the first time step $[0, 0, X_0]$ (illustrated in green in Fig.~\ref{fig:seq2col}), and the second time step $[0, 0, X_1]$ (illustrated in blue in Fig.~\ref{fig:seq2col}). Finally, these data blocks are concatenated into a one-dimensional continuous linear array, enabling CPU to execute LDRD and SMLAD instructions on the data structure. Additionally, the 2D input array of shape $(IC,K)$ is flattened into shape $(IC \cdot K, 1)$ to prepare for the Activation-Weight-Multiplication in next step.

We break down the multiplication of the $(L, IC, K)$ activation tensor and the $(OC, IC, K)$ weight tensor into three nested for-loops: the outermost loop iterates over L, the second over OC, and the innermost over the dot product between ${\text{input}}_{( IC \times K)}$ and ${\text{weight}}_{( IC \times K)}$. 

The memory access of input data and weight data is continuous and can be accelerated with the dual-MAC instructions like SMLAD or SMLALD. 


\subsection{Requantization and Fixed-Point Approximation}
\label{sec:requantization_fixed_point}
During the execution of CNNs, MAC operations have results of higher bit-width than their inputs, as previously discussed in Sec.~\ref{sec:Re-quantization}.
To convert the high-bit-width intermediate result into input activations for the next layer, a requantization process is required.

In principle, re-quantization would require a multiplication with a conversion factor in floating point format. However, such floating point operations would slow down the process severely on an MCU.
\begin{figure}
\includegraphics[width=8cm]{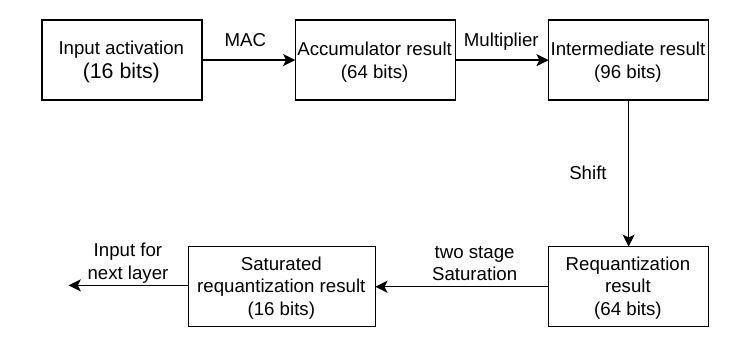} 
\centering
\caption{An overview of the requantization process, starting from input activations of a CNN layer, the MAC operations corresponding to the convolution, multiplication with the scaling factor of the layer, bit-shifting, and two-stage saturation to obtain a 16-bit activation for the next layer.} 
\label{fig:requantization}
\end{figure} 
Therefore, we propose a fixed-point approximation for our 16-bit quantization approach. 
As discussed in Sec.~\ref{sec:fixed_point}, the floating-point multiplication can be approximated by an integer multiplication and bit shifting. Fig.~\ref{fig:requantization} provides an overview of the requantization pipeline.


When a convolutional layer receives 16-bit input data, MAC operations with the weights result in 64-bit outputs: 32 bits to account for the maximum result of a 16-bit times 16-bit multiplication, and another 32 bits to leave enough space for accumulations. 

Then, the actual re-quantization begins. First, the 64-bit MAC output is multiplied with a 32-bit integer scaling factor, resulting in a 96-bit intermediate result.

We note that similar problems may occur for prior approaches, as well. For example, in the 
W8A16 quantization scheme from the CMSIS-NN library, MAC operations across more than 512 operands may result in a number exceeding 32 bits. After multiplication with a 32-bit scaling factor for re-quantization, the result may exceed 64-bits, yielding an overflow even if the result is stored as a long long type.
To avoid this problem, prior work applies the \texttt{arm\_nn\_requantize\_s64} function, which approximates the multiplication with the scaling factor with a 16-bit fixed-point operation:
$$S \approx  M_{reduced}\times 2^{-15-n}$$
that uses the same scheme as the fixed-point approximation described in Sec.\ref{sec:Re-quantization} but with 16 bits instead of 32 bits. 
However, compared to 32-bit multipliers, $M_{reduced}$ offers lower approximation accuracy for the scaling factor, resulting in undesirable quantization noise and loss of precision. Further, for W16A16 quantization, even the multiplication with a reduced 16-bit multiplier may lead to overflows beyond 64 bits. Hence, 
%
%
%
%
we propose a Full-Precision 96-bit Cross-Register Shift. 
Unfortunately, the standard C language (C99 and C11) does not support 96-bit data types (the longest integer type is long long with 64 bit); nor does it support a 96 bit cross-register shift.
Accordingly, a custom solution is required.

\begin{figure}
\includegraphics[width=8cm]{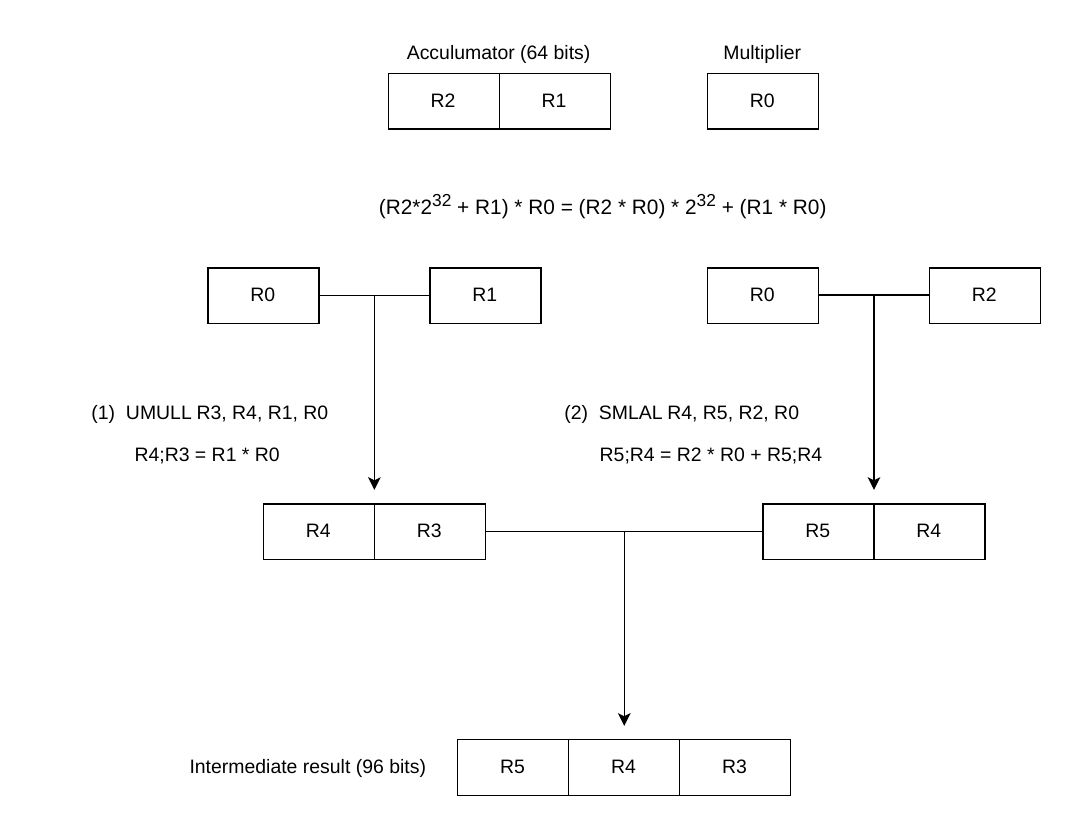} 
\centering
\caption{The multiplication of a 64-bit intermediate MAC result (accumulator) with a 32-bit multiplier for requantization using one intermediate UMULL for the lower bits and and one SMLAL operation for the higher bits, resulting in 94 bits spread across three registers, R3 for the low bits, R4 for the intermediate bits, and R5 for the high bits.} 
\label{fig:acc_mult}
\end{figure}

To bypass the limitations of the C language regarding 96-bit data, we directly cascade three 32-bit registers (as shown in Fig.\ref{fig:acc_mult}, R3: Low, R4: Mid, R5: High) at the assembly level into a 96-bit Data pipeline, accommodating a 32-bit multiplier without precision loss.

As shown in the Fig.~\ref{fig:acc_mult}, the data in the accumulator is 64 bits spread across a high-bit and low-bit register, with unsigned data in the low-bit register.

To implement the multiplication of the 64-bit accumulator with the 32-bit multiplier for requantization, we use the UMULL instruction to perform an unsigned multiplication of the low-order register R0 and the multiplier R1. The result is stored in R4 (middle) and R3 (low). Then, we use the signed-MAC instruction SMLAL to multiply the high 32-bit accumulator R2 by the multiplier R0 and directly accumulate the result into the register R5 (high) and R4 (middle). Finally, the result is stored in three registers, R5, R4 and R3 as a 96-bit number without precision loss.

After we obtain the full-precision 96-bit intermediate result, we perform a bit shift operation to finish the conversion into the quantization range of the next layer (refer to Fig.~\ref{fig:requantization}, top right). When the shifting amount is $Shift>32$, the low-order result in R3 is abandoned because it is shifted across the decimal point and would be lost after rounding. If the shift amount satisfies $0<Shift<32$, the low-order register is kept.  The bits shifted out of the middle-order register are merged into the low-order register.
 
\begin{figure}
\includegraphics[width=8cm]{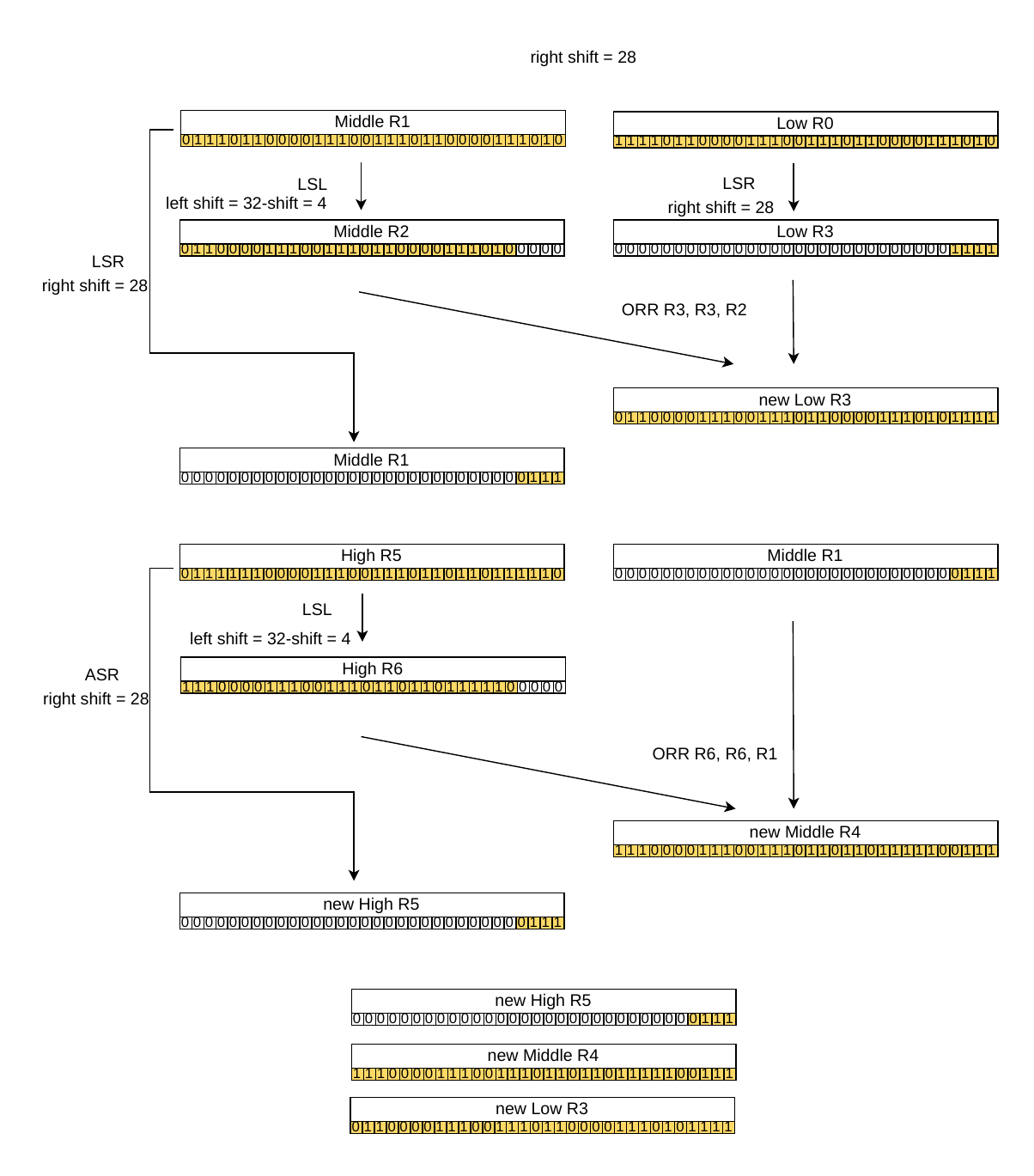} 
\centering
\caption{Full-Precision 96-bit Cross-Register Shift} 
\label{fig:96bits}
\end{figure}

More precisely, for shifts in the range $0<Shift<32$, the Full-Precision 96-bit Cross Register Shift can be defined as the following function (refer also to Fig.~\ref{fig:96bits}):
\begin{align*}
High_{new} &= High_{old} \gg_{ASR} Shift \\
Mid_{new} &= (Mid_{old} \gg_{LSR} Shift) \, \vert \, (High_{old} \ll (32 - Shift)) \\
Low_{new} &= (Low_{old} \gg_{LSR} Shift) \, \vert \, (Mid_{old} \ll (32 - Shift))
\end{align*}
For the low-order register, a LSR (Logical Shift Right) operation is performed by $Shift$ bits, with the vacated high-order bits filled with 0. Then, the middle-order register performs an LSL (Logical Shift Left) by $32-Shift$ bits. Finally, as Fig.~\ref{fig:96bits} shows, the two transformed values are combined using an ORR instruction to form a new 32-bit low-order result, which contains the data shifted from middle-order register. Then, the high-order register also performs an LSL by $32-Shift$ bits, and the middle-order register receives $32-Shift$ bits of data from the high-order register. Finally, the high-order register performs an ASR (Arithmetic Shift Right) instruction shifting by $32-Shift$ bits with sign-extension.

\subsection{Two-Stage Saturation}
\label{sec:two_stage_saturation}

In Sec.~\ref{sec:requantization_fixed_point}, we employ three registers to represent the 96-bit intermediate results of $Accumulator \times Multiplier$. Subsequently, we design a Full-Precision 96-bit Cross-Register Shift to complete the re-quantization operation.


To perform re-quantization, we continue to use a fixed point approximation scheme of the following form:
$$Out_{int} = (Accumulator \times Multiplier) >> Shift$$
As previous explained, Multiplier is an integer number (1 sign bit + 31 data bits), and shifting is normally above 30. The combination of these two operations approximates the multiplication with a floating-point scaling factor (e.g.\ 0.1).

To construct a larger example, suppose that after the $Accumulator \times Multiplier$ operation, the intermediate result is stored within three 32-bit registers (so 96 bits in total). The valid data of the intermediate result spans bits 0 through 45, while bits 46 through 95 are sign-extension. Assume we need to shift by 34 bits (e.g.\ to represent the multipliers 0.1). Then, the valid data stored at bits 34 through 45 is safely shifted down to bits 0 through 11. The discarded lower 34 bits of data represent actually the fractional part of the floating-point approximation. 
The data in bits 0 through 11 are the valid value required for the input of next layer.

Depending on the number of remaining bits after shifting, we designed three different processing schemes:
\paragraph{No saturation required (normal range):}
If the data after shifting $Result_{shifted}$ falls within 16 bits (1 sign-bit and 15 data-bit) range, $$-2^{15} \le x_{shifted} \le 2^{15} - 1$$, we use the data in low-order register directly as the input of the next convolutional layer.

\paragraph{16-bit saturation (normal overflow):}
If the shifted result exceeds the 16-bit representation limit, but the data still falls within 16-bit through 31-bit, i.e.
$$-2^{31} \le x_{shifted} < -2^{15} \quad \text{or} \quad 2^{15} - 1 < x_{shifted} \le 2^{31} - 1$$,
we utilize the assembly instruction SSAT. This operation saturates a 32-bit integer to 16 bits, namely to the range between 0x7FFF (32767) and 0x8000 (-32768). The 16-bit saturation can be defined as follows:
$$y_{16} = \min(\max(Result_{shifted}, -2^{15}), 2^{15} - 1)$$

\paragraph{Two-Stage Saturation (extreme overflow):}
In extreme cases, the data after shifting may exceed the representation limit of 32-bit, i.e.
$$x_{shifted} < -2^{31} \quad \text{or} \quad x_{shifted} > 2^{31} - 1 .$$
If we were to directly process bits 0 through 31 bits, the sign-bit in the higher registers would be ignored, yielding extreme quantization errors. For example, $2^{31}$ (0x80000000) is a positive number that exceeds the maximum representable range ($2^{31}-1$) of a 32-bit signed register. Hence, it is falsely interpreted as $-1$ in the 32-bit signed register. To avoid this mistake, we should first truncate $2^{31}$ to $2^{31} - 1$, which is the maximum value for a 32-bit signed register.

Accordingly, we propose a two stage Saturation scheme: we first saturate the data to the representation limit of 32-bit $$y_{32} = \min(\max(x_{shifted}, -2^{31}), 2^{31} - 1)$$, then saturate it to 16-bit using the assembly instruction SSAT. $$y_{16} = \min(\max(y_{32}, -2^{15}), 2^{15} - 1)$$.

After this process, we obtain the result for next convolutional layer and avoid the overflow risk.

Importantly, note that the saturation operations can implement a rectified linear unit (ReLU) transition function for a convolutional neural network by emplyoing the USAT operation instead of the SSAT operation.

\section{Evaluation results}

In our experiments, we compare three quantization approaches, W8A8, W8A16, and W16A16 (our proposed scheme) in terms of performance in machine learning tasks as well as execution time and energy consumption. The implementation of our experiment is available at: 
\url{https://gitlab.com/fastconv1d/tcn_experiments}


\subsection{Hardware and Configuration}
To comprehensively evaluate the latency, energy consumption and cross-platform versatility of our proposed quantization methods, we deployed our model and baselines on two representative MCUs.
\begin{itemize}
\item STM32H723ZG: ARM Cortex-M7 core with a maximum frequency of 550MHz, integrating with I-Cache and D-Cache with FPU.
\item GD32F450VET: ARM Cortex-M4 core with a maximum frequency of 200MHz, supported by FPU.
\end{itemize}
Both MCUs support the ARMv7E-M architecture and DSP extension instruction sets.

\paragraph{Quantization settings}

All machine learning models were trained using floating point numbers and then quantized after training (a so-called post training quantization). The W8A8 quantization was implemented using the official Pytorch.tq toolbox. However, this toolbox does not support W8A16 or W16A16 quantization. Consequently, we developed custom quantization approaches. During the calibration phase, we scaled the maximum observed activation values by a factor of 2.0 to prevent the precision loss due to truncation of unexpected activation outliers.

\paragraph{Compiler settings}
All C/Assembly inference operators are compiled using Arm Compiler 6, with -Ofast and link-time optimization (-flto). To measure the inference cycles precisely, we insert memory barrier instructions to eliminate the influence of global optimization on the measurement results.

\subsection{Regression}
To evaluate the regression performance and quantization noise of different quantization methods, we utilize the C-MAPSS Jet Engine Dataset and NinaPro DB2 dataset to evaluate the regression performance of different approaches. 
While a typical machine learning evaluation would focus on the regression error against the ground truth, our focus is on comparing different quantization schemes. Therefore, we take the output of the float32 TCN as reference and compute the root mean square error of the predictions of the quantized model against this reference.
$$RMSE_{quant} = \sqrt{\frac{1}{N}\sum_{i=1}^{N} \left( y_{FP32}^{(i)} - y_{Quantized}^{(i)} \right)^2}$$
This isolates the error due to quantization from the error due to model capacity.

\subsubsection{C-MAPSS Dataset}
\label{sec:c_mapss}
The C-MAPSS FD001 sub-dataset is provided by NASA, which is a standard benchmark for Remaining Useful Life (RUL) estimation of engines \citep{asif2022deep}. During flight, it is impossible to upload massive amounts of time-series data to the cloud, such that processing needs to occur at the engine itself using an MCU.
We use this dataset to evaluate the quantization errors of different quantization approaches.

The dataset consists of 100 train engines and 100 test engines with 26 input channels. To prevent early-life overestimation when the engine still works in a healthy state, we set a upper-bound threshold of 125 flights. The input channels contains 12 uninformative features, therefore, we preserve 14 input channels for our regression task \citep{asif2022deep}.
The window size of input data is set as 30 time steps \citep{yu2021prediction}.

We utilized a TCN model as shown in Fig.~\ref{fig:TCN_block} to perform the regression task. The TCN consists of 4 stacked TCN Blocks with each convolutional layer followed by weight normalization and each ReLU followed by dropout (dropout rate 0.3). The channel size of each block is 32, with kernel size of 5 and dilation $2^i$ where $i$ is the index of the TCN block (higher dilation for higher layers).  The final layer is a $1 \time 1$ convolutional layer to produce the actual RUL prediction for the engine.

We trained the TCN for 50 epochs with an Adam optimizer and the mean square error loss function, a batch size of 64, and a learning rate of 0.001.

The RMSE of the resulting non-quantized model against the ground truth is 13.1977 which suggests reasonable model fit \citep{wang2018remaining}. The non-quantized, float32 TCN serves as reference for the subsequent analysis of quantization errors.


\begin{table}[htbp]
\centering
\caption{Layer-wise, block-wise and End-to-End Quantization Error against FP32 Baseline on the C-MAPSS Jet Engine Dataset.}
\label{tab:quantization_rmse}
\begin{tabular}{lcccc}
\toprule
\textbf{Precision RMSE} & \textbf{Layer 0} & \textbf{Block 0} & \textbf{End-to-End} \\
\midrule
W8A8   & 0.01197 & 0.02456 & 1.89373 \\
W8A16  & 0.00194 & 0.00251 & 0.31093 \\
W16A16 & 0.00005 & 0.00012 & 0.01101  \\
\bottomrule
\end{tabular}
\end{table}

Tab.~\ref{tab:quantization_rmse} presents the RMSE of various quantization approaches relative to the float32 reference TCN model both for the full network (end-to-end) as well as the first layer and the first TCN block. A key phenomenon is that the quantization error increases with the network depth. For W8A8, the RSME of the first shallow layer is only 0.01197. However, this quantization noise is amplified through the cascade of causal dilated convolutional layers, raising to 0.02456 for the first TCN block and to 1.89373 for the full model. Note that this corresponds to roughly 14\% of the overall regression RMSE of the model, suggesting that the RMSE introduced by quantization alone is too high for practical purposes.

W8A16 increases the activation quantization resolution to 16-bit, which reduced the end-to-end RSME to 0.31093. This suggests that the resolution of the activations is the main factor driving high quantization RMSE. 
\begin{table}[htbp]
\centering
\caption{Quantization Error of the three quantization approaches against the float32 reference for one layer, one block, and the full TCN on the NinaPro DB2 force regression data set.}
\label{tab:ninaprodb2force}
\begin{tabular}{lcccc}
\toprule
\textbf{  }&\textbf{W16A16} & \textbf{W8A16} & \textbf{W8A8} \\
\midrule
RSME & 0.0015 & 0.0161 & 0.1970 \\
\bottomrule
\end{tabular}
\end{table}

Still, our proposed W16A16 scheme can reduce the end-to-end quantization RMSE by another order of magnitude to 0.01101, which is even similar to W8A8 in terms of the first shallow layer RSME.

\subsubsection{NinaPro DB2 force regression}
We utilize NinaPro DB2 as a second regression evaluation dataset \citep{gijsberts2014electromyography}. The task is to map surface electromyographic (sEMG) signals recorded at the lower arm of a person to the desired activation force of a hand prosthesis. This data set is particularly suitable for MCUs as the control of an arm prosthesis needs to be performed on the prosthesis itself using as little power and computation as possible \citep{islam2026development}.
Additionally, the force estimation via the regression model is very sensitive to the quantization noise. Therefore, this task serves as a useful benchmark for the evaluation of quantization approaches.

The NinaPro DB2 dataset contains data from 40 subjects, collecting sEMG signals via 12 channels at a sampling frequency at 2000Hz; the dataset includes 6 force signal channels, serving as targets for the regression task. 

The window size of our model is set to 300 samples; the window increment is set to 20 for the training set and 10 samples for the test set. Using finer increments during testing better simulates high-frequency mechanical control in the prosthesis.

Our regression model for NinaPro DB2 is similar to the model described in Sec.~\ref{sec:c_mapss}: it is also a TCN model composed of four TCN blocks and employs the same hyperparameters mentioned above, except that the input has 12 input channels and 6 output channels (corresponding to 12 sEMG channels and 6 force channels).

We use median value across 40 subjects to represent the RMSE of various quantization approaches, thereby avoiding the impact of outliers.

The median quantization RSME for W16A16 is 0.0015, whereas the RSME of W8A16 is 0.0161, and the RSME for W8A8 is 0.1970. Similar to the C-MAPSS data set, we observe an increase of ca.\ factor 10 from W16A16 to W8A16 and another factor of 10 from W8A16 to W8A8. The difference from W16A16 to the other methods is highly significant according to a Wilcoxon signed rank test ($p < 10^{-10}$).

\subsection{Classification}
In addition to regression tasks, we also evaluated the classification performance of various quantization approaches. As data set, we utilize NinaPro DB2, which contains 50 gestures for intent recognition, namely 17 hand and wrist movements, 23 grasping and functional movements, 9 force patterns and a rest position.
Given the high number of classes, we expect that data of different classes may be quite close to each other, meaning that quantization errors may be able to induce classification errors.

\begin{table}[htbp]
\caption{Classification accuracy on NinaproDB2 gestures based on sEMG signals.}
\label{tab:gesture_classification}
\begin{tabular*}{\textwidth}{@{} l c c c c @{}}
\toprule
\textbf{ } & \textbf{FP32} & \textbf{W16A16} & \textbf{W8A16} & \textbf{W8A8} \\
\midrule
Accuracy & 53.31\% & 53.29\% & 53.29\% & 41.63\% \\

\bottomrule
\end{tabular*}
\end{table}

As Tab.~\ref{tab:gesture_classification} shows, the accuracy of W16A16 and W8A16 is the same; about 0.02\% lower than the float32 model. However, W8A8 performs significantly worse ($p < 10^{-10}$ in a Wilcoxon signed rank test) with an accuracy drop of 11.66\%. Apparently, for a classification task, 16-bit activations have sufficient resolution to achieve comparable performance to the full float32 TCN, whereas 8-bit activations significantly hurt accuracy.

\subsection{Inference speed}
To evaluate the computing efficiency and computational throughput of our proposed framework and baselines, we benchmark on both the STM32H723ZG (ARM Cortex-M7) and GD32F450VET (ARM Cortex-M4) platforms.
The evaluation is implemented via two levels: a single layer and the full TCN model. 
The performance is quantified with CPU execution cycles.

\subsubsection{Layer-level benchmarking}
In this section, we selected the conv1d layers applied in TCN Regression model (refer to Tab.~\ref{tab:layer_benchmark_STM32H7} for the STM32H7 results and to Tab.~\ref{tab:layer_benchmark_GD32F4}). We observe that the W8A8 quantization achieves the lowest latency, followed by W16A16, while mixed-precision W8A16 exhibits the largest latency. This pattern repeats across various layer configurations and across platforms. Furthermore, GD32F4 needs roughly twice the execution cycles compared to STM32H7 due to the lack of I-Cache and D-Cache.

\begin{table}[htbp]
\caption{Layer-level benchmarking cycles comparison across different quantization precisions STM32H7 ($L=30$).}\label{tab:layer_benchmark_STM32H7}
\begin{tabular*}{\textwidth}{@{} l c c c  @{}}
\toprule
\textbf{Configuration} & \textbf{W8A8 } & \textbf{W8A16  } & \textbf{W16A16  } \\
\midrule
$14 \times 32 \times 5, D=1$ & $99867$  & $123257$  & $99436$  \\
$32 \times 32 \times 5, D=1$ & $144273$  & $194531$  & $164665$  \\
$14 \times 32 \times 1, D=1$ & $45861$  & $53789$  & $61374$  \\
$32 \times 1 \times 1, D=1$ & $5774$  & $6570$  & $6206$  \\
$32 \times 32 \times 5, D=2$ & $121790$  & $164095$  & $139373$  \\
$32 \times 32 \times 5, D=4$ & $77684$  & $104667$  & $88741$  \\
\bottomrule
\end{tabular*}
\end{table}

\begin{table}[htbp]
\caption{Layer-level benchmarking cycles comparison across different quantization precisions on GD32F4 ($L=30$).}\label{tab:layer_benchmark_GD32F4}
\begin{tabular*}{\textwidth}{@{} l c c c  @{}}
\toprule
\textbf{Configuration} & \textbf{W8A8 } & \textbf{W8A16  } & \textbf{W16A16  } \\
\midrule
$14 \times 32 \times 5, D=1$ & $172195$  & $222678$  & $194077$  \\
$32 \times 32 \times 5, D=1$ & $304884$  & $412798$  & $349351$  \\
$14 \times 32 \times 1, D=1$ & $82826$  & $103830$  & $109196$  \\
$32 \times 1 \times 1, D=1$ & $10363$  & $11816$  & $11121$  \\
$32 \times 32 \times 5, D=2$ & $272183$  & $349393$  & $295622$  \\
$32 \times 32 \times 5, D=4$ & $173375$  & $222449$  & $188165$  \\
\bottomrule
\end{tabular*}
\end{table}

\subsubsection{End-to-end model level benchmarking}
\label{ref:end_to_end}
\begin{figure}
\includegraphics[width=8cm]{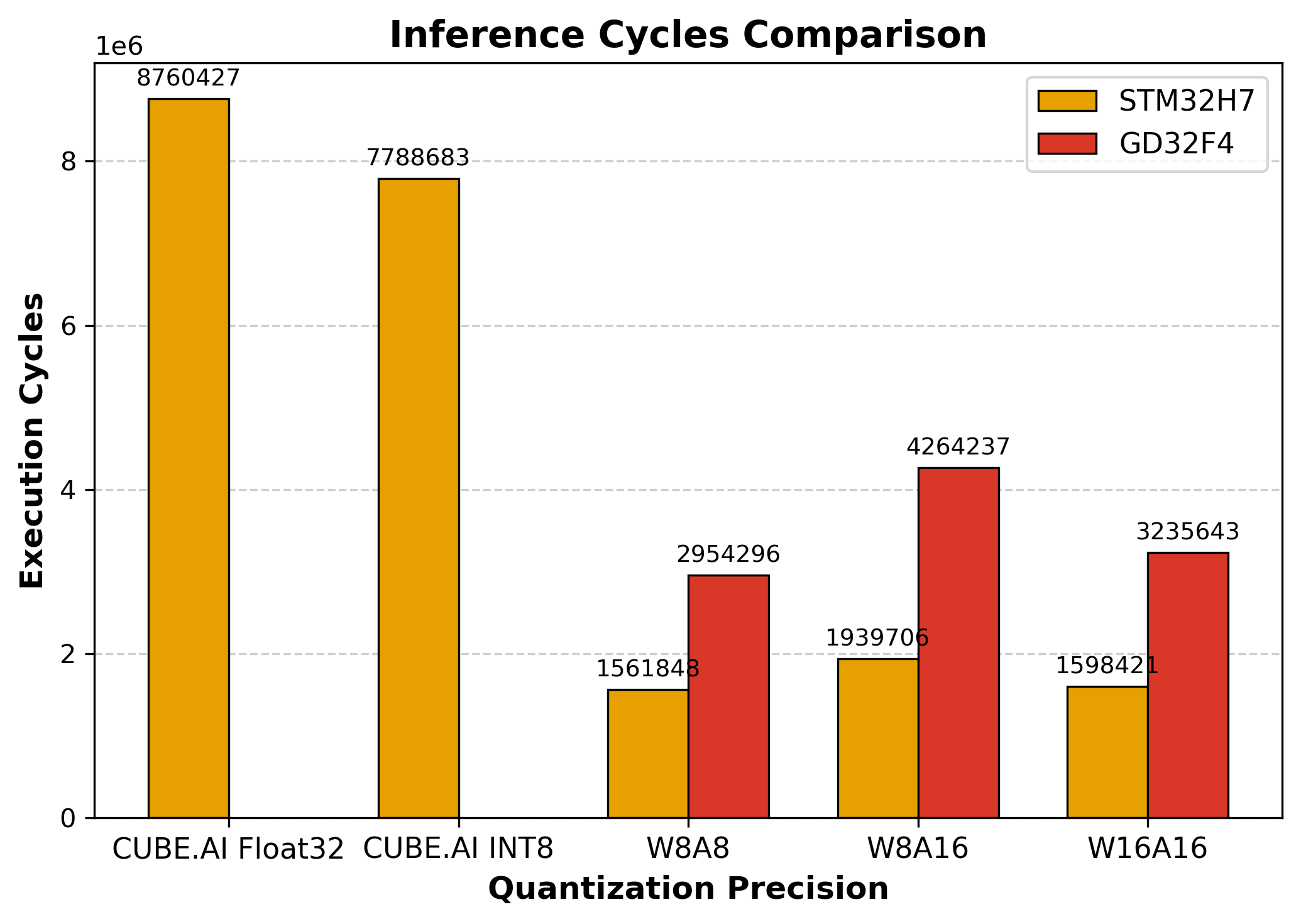} 
\centering
\caption{model level benchmarking} 
\label{fig:model}
\end{figure}

Figure~\ref{fig:model} shows the total execution cycles of the entire TCN regression model. Overall, we observe the same pattern as in the layer-level benchmark: W8A8 quantization is the fastest, followed by W16A16, with W8A16 being the slowest.
On the STM32H7 MCU, the W8A8 method achieves 2.29\% reduced cycles compared to W16A16. Our proposed W16A16 reduces cycles by 17.6\% compared to W8A16.
On the GD32F450 MCU, the difference is more striking: W8A8 saves 8.70\% cycles compared to W16A16 and W16A16 saves 24.1\% cycles compared to W8A16.

Our results indicate that W16A16 is faster than W8A16 on both the layer and the model level, providing both better precision and faster speed. While W8A8 is even faster, the impact of quantization errors in regression and classification tasks are likely not worth the time savings.

For the end-to-end benchmark, we also consider Cube.AI as an alternative scheme to execute TCNs on MCUs. However, Cube.AI was unable to convert some of the operations needed for the TCN layer (in particular left asymmetric padding) into 8-bit quantization. Accordingly, 77.5\% operations are floating point operations, severely impacting the speed of executing the TCN. Indeed, the Cube.AI INT8 and Float32 versions of the TCN perform similar on the STM32H7 hardware and are much slower than the other quantization approaches we tested. GD32F4 does not support Cube.AI, therefore, there is no results of Cube.AI on GD32F4. 

\subsection{Energy Consumption}
We measured the energy consumption with the Nordic Semiconductor's Power Profiler Kit II (PPK2), which is a precision energy analyzer. PPK2 is set as Source Meter providing the MCU with a 3.3V DC power supply.

To capture the time window for a single end-to-end neural network inference, we pulled a GPIO pin of MCU high once entering the neural network inference function and pulled it low when the inference is finished. This GPIO pin is connected to the Logic Port of PPK2 to achieve a hardware synchronization between the inference process and the transient current waveform.

The sampling frequency is set as 100,000 sample points per second, within a inference window the accumulated charge $Q_{\text{total}}$, the inference time $T_{inference}$ and average current amplitude are measured $I_{\text{average}}$. 
The energy per inference is proportional via the voltage: $E_{\text{inf}} = 3.3V \cdot Q_{\text{total}} = 3.3V\cdot I_{average}\cdot T_{inference}$.

\begin{table}[htbp]
\caption{Inference time and Energy Characterization on STM32H7.}
\label{tbl_power_energy_benchmark}
\footnotesize
\begin{tabular*}{\textwidth}{@{} l c c c c c c @{}}
\toprule
\textbf{Metric} & \textbf{Sleep} & \textbf{Cb.32}& \textbf{Cb.8}& \textbf{W8A8} & \textbf{W8A16} & \textbf{W16A16} \\
\midrule
Avg. I (mA) & 53.33 & 189.30 & 178.34 & 197.14 & 208.15 & 212.06 \\
Inference Time (ms)  & --    & 15.11 & 13.68 & 2.77 & 3.42 & 2.82 \\
Total Energy (mJ)    & --    & 9.44 & 8.05 & 1.80 & 2.35 & 1.95 \\
\bottomrule
\end{tabular*}
\end{table}

\begin{table}[htbp]
\caption{Inference time and Energy Characterization on GD32F4.}
\label{tbl_power_energy_benchmark_GD32F4}
\begin{tabular*}{\textwidth}{@{} l c c c c @{}}
\toprule
\textbf{Metric} & \textbf{Sleep} & \textbf{W8A8} & \textbf{W8A16} & \textbf{W16A16} \\
\midrule
Avg. I (mA) & 85.96 & 122.43 & 124.36 & 120.74 \\
Inference Time (ms)  & -- & 14.46 & 20.59 & 15.90 \\
Total Energy (mJ)    & -- & 5.84 & 8.45 & 6.34 \\
\bottomrule
\end{tabular*}
\end{table}


Table~\ref{tbl_power_energy_benchmark} shows the measurement results for the STM32H7 platform. Overall, the same pattern repeats again: W8A8 needs the least energy and time, followed by W16A16 and W8A16 needs the most energy and time. In more detail, the full TCN model on STM32H7 is 0.6 ms faster and uses 0.4 mJ less energy for W16A16 quantization compared to W8A16 quantization. W8A8 is only 0.05 ms faster than W16A16 and needs 0.15 mJ less energy.

Cb.32 denotes Cube.AI float32 and Cb.8 means Cube.AI INT8. As shown in Tab.~\ref{tbl_power_energy_benchmark}, both are slow (ca.\ 5x compared to W16A16) and require high energy (ca.\ 4x compared to W16A16). 

For the GD32F4 platform (Tab.~\ref{tbl_power_energy_benchmark_GD32F4}), the pattern repeats once more but is more prominent. W8A16 needs 4.69 ms more time and 2.11 mJ more energy compared to W16A16; and W16A16 is 1.44 ms slower and needs 0.5 mJ more energy compared to W8A8.

Sleep denotes the current of sleep mode. By subtracting this value, we can calculate the 
the algorithm's actual energy consumption, namely, "Algorithmic Execution Energy", which contains the CPU computing consumption and memory access consumption. It can be computed via:
$E_{\text{execution}} = 3.3V\cdot (I_{argorithmic}-I_{sleep})\cdot T_{inference}$.
For example, the Algorithmic Execution Energy of W16A16 on STM32H7 can be computed as: $E_{\text{execution}} = 3.3V\cdot (I_{W16A16}-I_{sleep})\cdot T_{inference}= 3.3V\cdot (212.06mA-53.33mA)\cdot 2.82ms = 1.47mJ$


\section{Conclusion}

In this paper, we propose a W16A16 quantization scheme to execute deep neural networks, in particular time convolutional neural networks on MCUs. We provided the exact assembly instructions needed to run W16A16 quantization efficiently, implemented the quantization scheme as an open source package, and performed experiments to compare W16A16 quantization to W8A8 and W8A16 quantization across two hardware platforms and three machine learning tasks. Our results challenge the notion that lower quantization precision automatically leads to faster speed. Our W16A16 quantization method is both faster (ca.\ 17.6\% for a full TCN model on STM32H7) and more energy-efficient than W8A16 across MCUs, at both layer-wise and end-to-end model level, while providing higher accuracy in both regression and classification tasks. 
While W8A8 is slightly faster and more energy efficient compared to our proposed W16A16 scheme, the empiric classification and regression performance of the quantized neural net is significantly worse (ca.\ 100x quantization error in regression tasks). Therefore, W16A16 provides an attractive option for applications with high accuracy requirements under tight energy and time constraints.

This work also provides the explanation for the empiric observations based on the underlying architecture: in the Cortex-M instruction set, the maximum throughput for integer MAC operations is the same for 8-bit and 16-bit operations and 8-bit operations require additional pre- and post-processing instructions, thus yielding overhead for W8A16 quantization.
Based on our architecture analysis, we modified the 64-bit requantization provided by CMSIS-NN, and designed a Full-Precision 96-bit Cross-Register Shift and Two-Stage saturation to avoid the overflow of intermediate values.

Further work could explore whether such architecture-specific resolution gains with little energy/time overhead can also be translated to other MCUs or hardware architectures.

\subsection{Limitations}
The cost advantage of our proposed W16A16 quantization method is based on the architecture of ARMv7E-M which exhibits the same speed for 8-bit and 16-bit operations. However, this special effect does not necessarily translate to newer architectures such as ARMv8.1-M or RISC-V, which are capable to process four 8-bit data elements in a single instruction of MAC operation. Still, ARMv7E-M is a widely applied cortex-M architecture, providing a wide scope of application of our approach at this time.

We also note that our experiments are limited to TCN models and three machine learning tasks across two data sets. On other data sets, other tasks, and other models, results may be slightly different. Still, given the consistent patterns we observed across metrics, data sets, and hardware in our experiments, we are fairly confident that the qualitative pattern would generalize: W16A16 is faster than W8A16 and roughly comparable to W8A8, while W8A8 is significantly less precise compared to the other two.



\bibliographystyle{plainnat}
\bibliography{literature}
\end{document}